\documentclass[10pt,twocolumn,letterpaper]{article}

\usepackage{iccv}
\usepackage{times}
\usepackage{epsfig}
\usepackage{graphicx}
\usepackage{amsmath}
\usepackage{amssymb}
\usepackage{booktabs}
\usepackage{rotating}
\usepackage{footnote}

\usepackage{tablefootnote}
\usepackage{threeparttable}

\usepackage[pagebackref=true,breaklinks=true,letterpaper=true,colorlinks,bookmarks=false]{hyperref}

\iccvfinalcopy 

\def\iccvPaperID{} 

\usepackage{multirow}

\usepackage{enumitem}

\usepackage{makecell}
\usepackage{array}
\usepackage{algorithm}
\usepackage{algorithmic}
\usepackage[table,xcdraw]{xcolor}
\usepackage[accsupp]{axessibility}  

\usepackage{colortbl}
\usepackage{hhline}
\usepackage{subfig}
\usepackage[capitalize]{cleveref}
\crefname{figure}{Fig.}{Figs.}
\crefname{section}{Sec.}{Secs.}
\Crefname{section}{Section}{Sections}
\Crefname{table}{Table}{Tables}
\crefname{table}{Tab.}{Tabs.}

\ificcvfinal\fi

\begin{document}

\title{Attentive Mask CLIP}


\author{Yifan~Yang$^1$\footnotemark[1] 
\quad Weiquan~Huang$^{2 \dag}$\thanks{Equal contribution. \dag This work was done during internship in MSRA.}
\quad Yixuan~Wei$^{3 \dag}$
\quad Houwen~Peng$^1$ 
\quad Xinyang~Jiang$^1$ \\
\quad Huiqiang~Jiang$^1$ 
\quad Fangyun~Wei$^1$ 
\quad Yin~Wang$^2$ 
\quad Han~Hu$^1$
\quad Lili~Qiu$^1$
\quad Yuqing~Yang$^1$ \\
{$^1$Microsoft Research Asia}  \quad {$^2$Tongji University} \quad {$^3$Tsinghua university}\\
\small{
\texttt{\{yifanyang,houwen.peng,xinyangjiang,hjiang,fawe,hanhu,liliqiu,yuqing.yang\}@microsoft.com}}\\
\small{\texttt{\{weiquanh,yinw\}@tongji.edu.cn} \quad\small{\texttt{wei-yx20@mails.tsinghua.edu.cn} } }}

\maketitle
\ificcvfinal\thispagestyle{empty}\fi

\begin{abstract}
In vision-language modeling, image token removal is an efficient augmentation technique to reduce the cost of encoding image features. The CLIP-style models, however, have been found to be negatively impacted by this technique. 
We hypothesize that removing a large portion of image tokens may inadvertently destroy the semantic information associated to a given text description, resulting in misaligned paired data in CLIP training. 
To address this issue, we propose an attentive token removal approach, which retains a small number of tokens that have a strong semantic correlation to the corresponding text description. The correlation scores are dynamically evaluated through an EMA-updated vision encoder. Our method, termed attentive mask CLIP, outperforms original CLIP and CLIP variant with random token removal while saving the training time. In addition, our approach also enables efficient multi-view contrastive learning. 
Experimentally, by training ViT-B on YFCC-15M dataset, our approach achieves 43.9\% top-1 accuracy on ImageNet-1K zero-shot classification, 62.7/42.1 and 38.0/23.2 I2T/T2I retrieval accuracy on Flickr30K and MS COCO, outperforming SLIP by \textbf{+1.1\%}, \textbf{+5.5/+0.9}, and \textbf{+4.4/+1.3}, respectively, while being \textbf{2.30$\times$} faster. 
An efficient version of our approach runs \textbf{1.16$\times$} faster than the plain CLIP model, while achieving significant gains of \textbf{+5.3\%}, \textbf{+11.3/+8.0}, and \textbf{+9.5/+4.9} on these benchmarks, respectively. Code will be release in \url{https://github.com/microsoft/A-CLIP}.

\end{abstract}
\section{Introduction}

\begin{table}[]
\centering
\resizebox{\columnwidth}{!}{%
\begin{threeparttable}
\begin{tabular}{c|c|c|c|c|c} 
\toprule
\multirow{2}{*}{Methods}                 & \multirow{2}{*}{\begin{tabular}[c]{@{}c@{}}Training\\ Time\end{tabular}} & \multirow{2}{*}{\begin{tabular}[c]{@{}c@{}}GPU\\ Memory\end{tabular}} & IN 1K         & Flickr30K          & MS COCO             \\ 
\cline{4-6}
                                         &                                                                          &                                                                       & 0-shot        & I2T/T2I            & I2T/T2I             \\ 
\hline
CLIP                                     & 1.00×                                                                    & 14G                                                                   & 37.6          & 51.4/32.6          & 27.9/17.6           \\
SLIP                                     & 2.67×                                                                    & 30G                                                                   & 42.8          & 57.2/41.2          & 33.6/21.9           \\
MaskCLIP                                 & 1.56×                                                                    & 16G                                                                   & 42.7          & 60.0/38.8          & 34.1/21.2           \\ 
\hline
\rowcolor[rgb]{0.925,0.957,1} A-CLIP     & 1.16×                                                                    & 14G                                                                   & \textbf{43.9} & \textbf{62.7/42.1} & \textbf{38.0/23.2}  \\
\rowcolor[rgb]{0.925,0.957,1} A-CLIP-eff & \textbf{0.86×}                                                           & \textbf{13G}                                                                   & 42.9          & \textbf{62.7}/40.6          & 37.4/22.5           \\
\bottomrule
\end{tabular}
\begin{tablenotes}
\item[1] The full training wall clock time and GPU memory footprint are measured on the same device. We report the training cost relative to the original CLIP~\cite{AlecRadford2021LearningTV}.
\end{tablenotes}
\end{threeparttable}
}
\caption{We compare our attentive mask CLIP (A-CLIP) with CLIP~\cite{AlecRadford2021LearningTV}, SLIP~\cite{NormanMu2022SLIPSM} and MaskCLIP~\cite{XiaoyiDong2022MaskCLIPMS}. A-CLIP outperforms CLIP by +6.3\%, +11.3/+9.5 and +10.1/+5.6 on Imagenet-1K~\cite{OlgaRussakovsky2014ImageNetLS} zero-shot classification, Flickr30K~\cite{flickr30k} and MS COCO~\cite{TsungYiLin2014MicrosoftCC} I2T/T2I retrieval. An efficient variant termed A-CLIP-eff outperforms CLIP by +5.3\%, +11.3/+8.0, and +9.5/+4.9 on these benchmarks, while reducing the training time to $0.86\times$.}

\label{tab:main_results}
\end{table}

\begin{figure}[!ht] 
\centering 
\includegraphics[width=1.03\columnwidth]{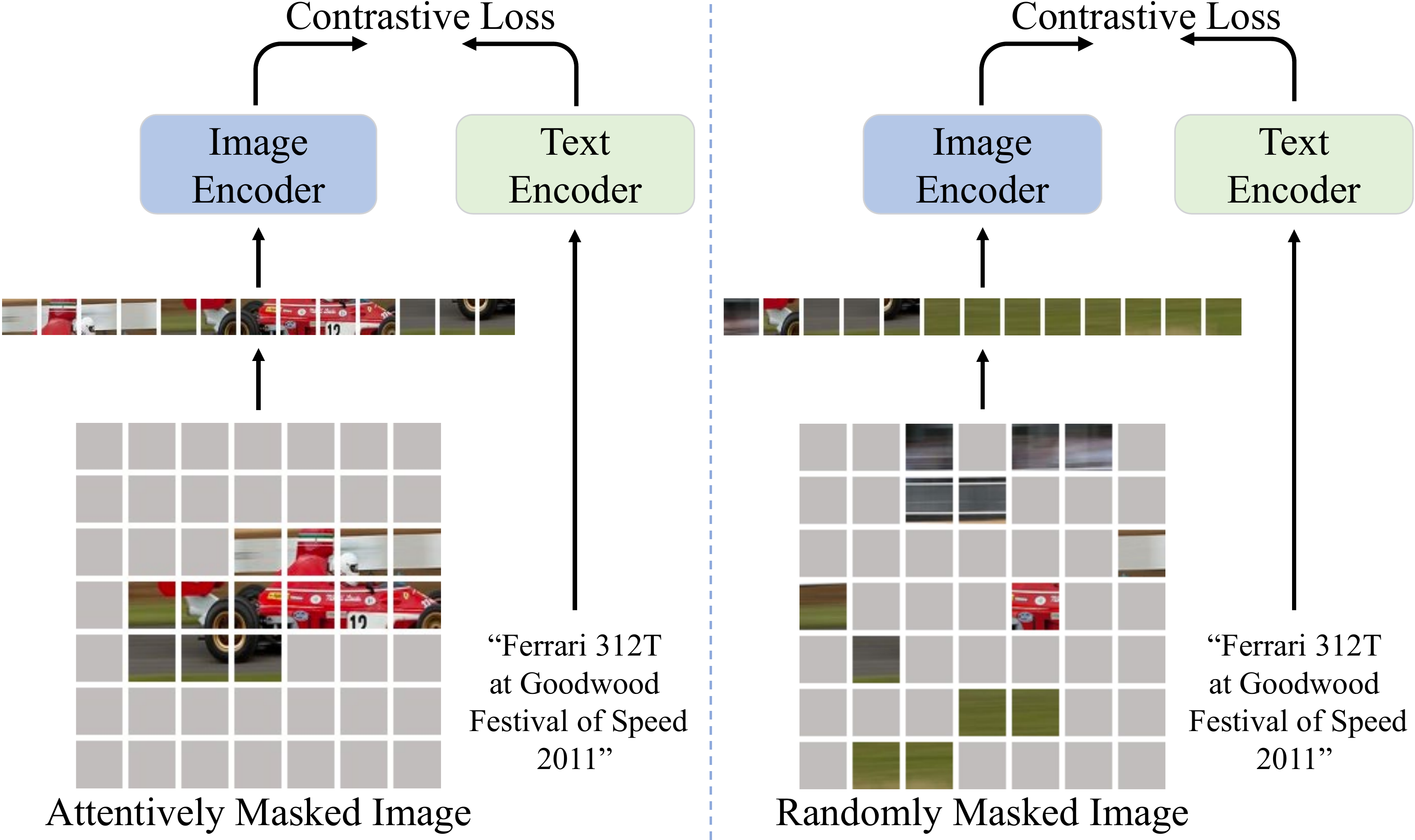} 
\caption{Attentive mask vs. random mask. Left part is the attentive mask applied CLIP training process, and right part is random mask applied. Here we use ViT-B16 model from A-CLIP's visual encoder to generate above masked images with patch size of $32 \times 32$ and $25\%$ mask ratio. The image and alt-text are sampled from YFCC-100M~\cite{BartThomee2022YFCC100MTN}.}
\label{figure:masking} 
\end{figure}

Large-scale vision-language pre-training models, such as CLIP \cite{AlecRadford2021LearningTV} and ALIGN \cite{ChaoJia2021ScalingUV}, have demonstrated remarkable capabilities in zero-shot image classification and multi-modal retrieval. However, these models typically require a large amount of training data, which raises the training cost. For instance, CLIP is trained on 400 million image-text pairs, and ALIGN uses more than 1 billion paired data. This raises the need for more efficient language-image pre-training methods.

This paper aims to improve the efficiency of CLIP training by introducing an efficient image augmentation approach called ``image token removal". This approach drops a large portion of image tokens, thereby reducing the computation in the image encoder. It has been shown to be effective in masked image modeling~\cite{KaimingHe2021MaskedAA,ZhirongWu2022ExtremeMF,ShlokMishra2022ASE,CMAE,SIM} when combined with vision Transformer architectures~\cite{Vit}. Thus, we seek to introduce this approach into CLIP training to improve its efficiency.

However, previous work has shown that dropping tokens randomly can harm CLIP performance \cite{ZiYiDou2021AnES,WonjaeKim2021ViLTVT}. This is also evidenced in our own experiments (see Table~\ref{tab:attn_ablation}). Specifically, we observe a -2.6\% top-1 accuracy drop on ImageNet zero-shot  classification when we randomly remove 50\% tokens in each image. The underlying reason for this issue is that the process of removing tokens may mistakenly eliminate semantic content that is pertinent to the alt-text, leading to inaccurate image-text pairs for CLIP training.

To mitigate this issue, we propose an attentive token removal strategy, as shown in Figure~\ref{figure:masking}. 
The fundamental idea is to retain a small set of image tokens that are more closely related to the corresponding text description while discarding those that are irrelevant. Concretely, we encode all image tokens into a latent space and then calculate the correlation scores between each image-token feature and the text feature extracted by the CLIP's text encoder. We investigate several strategies for selecting image tokens using correlation scores and conclude that retaining these image tokens that are most semantically related to the text description yields the best performance. The correlation scores are computed by using the exponential moving average (EMA) version of the vision encoder. Specifically, we use the attention weights of the $\left [ CLS \right ]$ token of the visual encoder as the correlation scores. We also find that using the averaged attention weights of all layers performs better. Figure~\ref{figure:vis} shows the selected tokens, which correlate well with the associated text semantics.

The efficiency of the proposed token removal approach allows us to construct multiple masked views from an image while keeping the training as efficient as the original CLIP, e.g., 2 masked views with a token removal ratio of 50\%. The random cropping strategy adds certain stochastic effects to different masked views, enabling the application of an auxiliary contrastive loss between the augmented views in addition to the plain CLIP loss. In fact, as has been shown in SLIP~\cite{NormanMu2022SLIPSM,XiaoyiDong2022MaskCLIPMS}, the auxiliary task facilitates the CLIP training. We consider both SimCLR~\cite{Simclr} and SimSiam~\cite{XinleiChen2020ExploringSS} methods for the auxiliary contrastive learning task. It is worth noting that the EMA branch can be naturally treated as another view, and we also apply an online-to-EMA contrastive or consistency loss and use the formulation from BYOL~\cite{JeanBastienGrill2020BootstrapYO} for this purpose.

The proposed approach is called A-CLIP, which introduces only 16\% computational overhead in comparison to the plain CLIP. It is also 2.30$\times$ and 1.34$\times$ faster than previous CLIP improvements that also include multiple views and additional self-supervised losses, while being more effective. Using ViT-B and the YFCC-15M~\cite{BartThomee2022YFCC100MTN,AlecRadford2021LearningTV} dataset, the A-CLIP framework achieves 43.9\% top-1 accuracy on ImageNet-1K~\cite{OlgaRussakovsky2014ImageNetLS} zero-shot classification(see Table~\ref{tab:main_results}). Additionally, it achieves 62.7/42.1 and 38.0/23.2 I2T/T2I retrieval accuracy on Flickr30K~\cite{flickr30k} and MS COCO~\cite{TsungYiLin2014MicrosoftCC}, respectively, which is \textbf{+1.1\%}, \textbf{+5.5/+0.9}, and \textbf{+4.4/+1.3} higher than the SLIP method, and \textbf{+1.2\%}, \textbf{+2.7/+3.3}, and \textbf{+3.9/+2.0} higher than the MaskCLIP method.

Also note the training cost of our approach can be further reduced by using a lower resolution input for the EMA network, i.e., 2x lower reduced resolution. This has little affect on the accuracy of correlation score computing, while marginally reduce the efficacy of online-to-EMA contrastive loss. This strategy will reduce the EMA computation by more than 4$\times$, resulting in an efficient variant that is even 1.16$\times$ faster than the plain CLIP model, and is significantly more accurate. We refer to this more efficient variant as A-CLIP-eff.

\section{Related Work}

\paragraph{Contrastive language-image pre-training}
An important goal of computer vision is to interpret visual signals using language that humans can understand. While the field has long used image classification tasks to learn visual representations which connect visual signals to semantics, recent works, such as CLIP~\cite{AlecRadford2021LearningTV} and ALIGN~\cite{ChaoJia2021ScalingUV}, suggest a new way to connect visual signals with linguistic semantics by contrasting image-language pairs. In this visual-language contrastive learning framework, the training data is more scalable as billions of image-alt-text pairs can be easily collected from the internet. At present, CLIP has become a mainstream visual learning method that not only learns transferable representations but also connects visual signals to arbitrary semantics. There have been extensive follow-up studies~\cite{JunnanLi2022BLIPBL,YutingGao2022PyramidCLIPHF,LeweiYao2021FILIPFI,NormanMu2022SLIPSM,XiaoyiDong2022MaskCLIPMS,iCAR2022,UniCL2022} to improve the effectiveness by sacrificing training efficiency. Our method aims to improve CLIP pre-training while improving both effectiveness and efficiency.

\paragraph{Masking tokens for efficient computation}
 Vision Transformers~\cite{Vit,liu2021swin} process images as sequences of patch tokens and perform encoding computation on these tokens. A masking strategy that removes certain tokens from an image can significantly speed up the computation of the image encoder. Dynamic ViTs~\cite{rao2021dynamicvit,wang2021dynamic} learn to remove tokens for efficient image classification. Masked autoencoder~\cite{KaimingHe2021MaskedAA} randomly masks 75\% of tokens, which significantly speeds up self-supervised visual representation pre-training based on masked image modeling.

This paper extends the idea of token masking to speed up the CLIP training process. Instead of using random masking like masked image modeling, we propose an attentive masking method that removes only semantically meaningless tokens, thereby alleviating the issue of noisy pairing between image tokens and text descriptions in the random masking strategy caused by incorrectly discarded semantic content.

\paragraph{Comparison with a concurrent work FLIP~\cite{FLIP}}
Similar to our approach, there is concurrent work called FLIP~\cite{FLIP}, which also employs token masking to accelerate CLIP training. However, FLIP uses a random masking strategy like MAE~\cite{KaimingHe2021MaskedAA}, and achieves inferior zero-shot accuracy than when trained with full images, while keeping the batch size and using the ViT-B model.

Our approach improves upon FLIP in several ways. First, we propose an attentive masking strategy for CLIP training that significantly outperforms the random masking baseline. This is especially crucial for CLIP, as its target heavily relies on semantic texts, while random masking can work well with MIM that do not have explicit semantic supervision. Second, we introduce multiple masked image views into our framework, enabling us to conveniently incorporate auxiliary pre-text tasks, such as image-to-image contrast learning. By reducing computation through masking, our approach does not increase computation compared to the plain CLIP model using full images, while improving the pre-trained representations.

With these techniques, our approach achieves significant accuracy improvements over the original CLIP model on both zero-shot image classification and multi-modal retrieval. In contrast, FLIP only achieves comparable performance to the original CLIP model, even with additional training tricks. Also note that some of the findings in FLIP, particularly on scaling experiments and hyper-parameter tuning such as larger batch sizes and base learning rate tuning, complement to our approach. Together, these findings provide readers with a more complete view of using masking for CLIP training. 

\paragraph{Masking for data augmentation}
Token masking is not only more efficient, but also serves as a form of data augmentation for visual representation learning. Studies have shown that masking augmentation performs well in instance-based contrastive learning~\cite{MahmoudAssran2022MaskedSN,ZhirongWu2022ExtremeMF,LiJing2022MaskedSC,ShlokMishra2022ASE}. While not our primary focus, we demonstrate that attentive masking for CLIP training also benefits from its data augmentation property, as reflected in the increasing accuracy with a longer scheduler where saturation is observed using full images as in \Cref{tab:epochs}.

\paragraph{Combining CLIP with other representation learning methods}
Our method is also related to recent approaches that combine CLIP with other pre-text tasks to improve representation learning~\cite{NormanMu2022SLIPSM}. For example, SLIP~\cite{NormanMu2022SLIPSM} combines CLIP with image-to-image contrastive learning~\cite{moco,Simclr}, while MaskCLIP~\cite{XiaoyiDong2022MaskCLIPMS} combines CLIP with masked image modeling~\cite{HangboBao2021BEiTBP,KaimingHe2021MaskedAA,ZhendaXie2021SimMIMAS,swinv2}. These approaches have shown to learn stronger representations, but at the cost of higher training requirements.

In contrast, our method introduces the task of image-to-image contrastive learning without incurring additional training costs. The attentive mask generated by an EMA network on the full image constitutes another view of the image, enabling us to add an auxiliary task of image-to-image contrastive learning with minimal additional cost. Moreover, the masking strategy allows us to introduce more masked views with minimal computation cost, facilitating the application of richer image-to-image tasks in training and further improving representation learning. Our framework can also naturally incorporate masked image modeling, as we use a masked image input, which is a direction for future work.
  \section{Method}

\begin{figure}[!tb] 
\centering 
\includegraphics[width=0.47\textwidth]{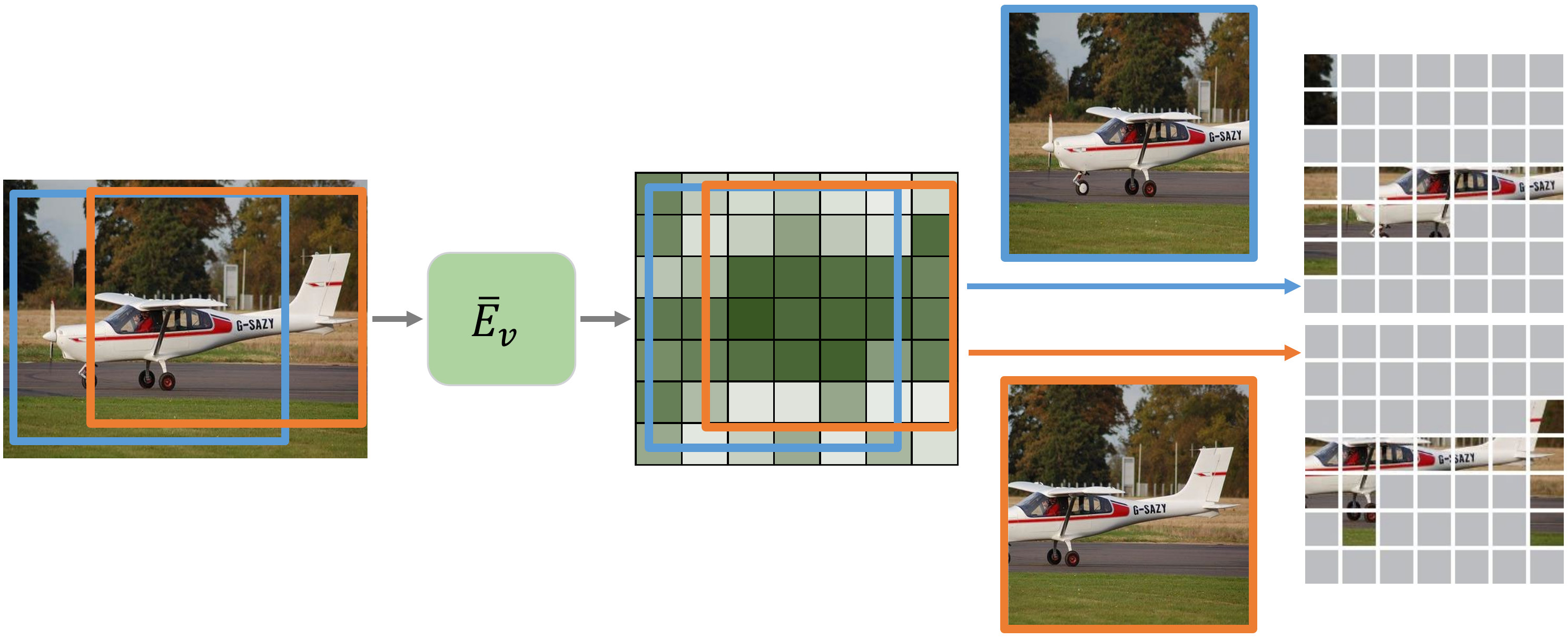} 
\caption{A illustration of computing attentive masks for multiple image views. For different image views, we perform score map computation once using an EMA version of the visual encoder ($\overline{E_v}$) on the original image or the minimum enclosing area of all image views. Then for each image view, a bilinear sampling approach is adopted to generate the selection score for each image token. The masking is performed based on the selection scores.}
\label{figure:attn} 
\end{figure}

\subsection{A Brief Review of CLIP}
CLIP~\cite{AlecRadford2021LearningTV} is a visual representation learning approach that uses a large amount of image-to-text pairs. The training task treats each image and its associated alt text as a positive pair and pairs of this image to all other alt-texts as negative. For all possible pairs within a batch, the method applies an InfoNCE-like loss to classify them as positive or negative. The CLIP model shows well connecting images to arbitrary language semantics and achieves remarkably strong zero-shot image recognition accuracy on benchmarks. Its learned representation also performs very well when fine-tuned on various down-stream tasks~\cite{li2021align}.

Specifically, the CLIP method independently applies a visual encoder $\mathbf{E}_v$ and a language encoder $\mathbf{E}_l$ on each image and each alt-text, respectively. In this paper, the visual and language encoders are instantiated as a Vision Transformer~\cite{Vit} and a standard language Transformer~\cite{devlin2018bert}. For the vision Transformer, a learnable $\left [ CLS \right ]$ token is prepended to the image tokens to represent the entire image. For the language encoder, an $\left [ EOS \right ]$ token is appended after the last word to represent the full sentence. The CLIP model projects the $\left [ CLS \right ]$ and $\left [ EOS \right ]$ features into an embedding space, denoted as $e^I$ and $e^T$ respectively, where a vision-language contrastive loss is applied:
\begin{equation}\label{eq:CLIPOut}
\mathcal L_{vl} = 0.5 * \mathcal L_v + 0.5 * \mathcal L_l
\end{equation}
where 
\begin{equation}\label{eq:CLIPILoss}
\mathcal L_v = -\frac{1}{B} \sum_{i=1}^B \log \frac{\exp \left(\operatorname{sim}\left(e_i^I, e_i^T\right) / \tau\right)}{\sum_{j=1}^B \exp \left(\operatorname{sim}\left(e_i^I, e_j^T\right) / \tau\right)}
\end{equation}
\begin{equation}\label{eq:CLIPTLoss}
\mathcal L_l = -\frac{1}{B} \sum_{i=1}^B \log \frac{\exp \left(\operatorname{sim}\left(e_i^T, e_i^I\right) / \tau\right)}{\sum_{j=1}^B \exp \left(\operatorname{sim}\left(e_i^T, e_j^I\right) / \tau\right)}
\end{equation}
In the above equations, $B$ denotes the batch size; $\operatorname{sim}(\cdot)$ denotes the cosine similarity function; $\tau$ is a learnable temperature to scale the logits.

\subsection{Masking for Efficient CLIP Training}

We seek to adopt a masking strategy to improve the training efficiency of CLIP.

A naive approach is to randomly select a portion of image tokens to remove, as shown in Figure~\ref{figure:masking}~(right). The random masking method has been shown to be effective for masked image modeling(MIM)~\cite{KaimingHe2021MaskedAA}. However, we note two main differences between MIM and CLIP tasks: 1) The MIM method is mainly designed for pre-training, while CLIP excels at zero-shot classification and retrieval. In fine-tuning settings, the domain gap between masked-image based pre-training and full-image based fine-tuning can be bridged by adjusting model parameters, which zero-shot evaluation hopes to achieve good accuracy by not changing the original model and thus requires a small domain gap between pre-training and evaluating; 2) MIM pre-training is more of a low-level task without language semantics. Removing highly semantic visuals does not affect training in general. CLIP training is more of a high-level task, aiming to connect images to semantics defined by text. Removing highly semantic objects related to associated alt-text can form noisy pairs. As illustrated in Figure~\ref{figure:masking}~(right), after the Ferrari car is almost discarded from an image, it would be inappropriate to associate the masked image to a textual description of Ferrari at a race. 

In fact, as shown in Table~\ref{tab:attn_ablation}, using a random ratio with 50\% mask ratio, the zero-shot accuracy on ImageNet-1K is 2.6\% worse than using the full image for training (35.0\% vs. 37.6\%). In the following subsection, we will present an attentive masking method that can resolve the issues of random masking described above, and achieves good accuracy for zero-shot and retrieval.

\subsection{Attentive Mask}
\label{sec:attn_mask}

The goal of attentive masking is to retain tokens that are relevant to language description, as illustrated in Figure~\ref{figure:masking}~(left).

To this end, we note that the representation of the $\left [ CLS \right ]$ token after CLIP training corresponds well to the semantics contained in the associated alt-text, and its attention weights to other image tokens can act a good indicator as the relevance measure of each image token to the linguistic semantics. We thus compute the score of the token at location $P$ as:
\begin{equation}\label{eq:Attn}
s_P
=\frac{1}{HL} \sum_{l=1}^{L} \sum_{h=1}^{H} \operatorname{Softmax}\left(\frac{\mathbf{f}^q_{lh}({CLS}) \cdot {{\mathbf{f}^k_{lh}(P)}}}{\sqrt{C}}\right),
\end{equation}
where $l$ denotes the layer index; $h$ denotes the attention head index; $\mathbf{f}_{lh}^q(CLS)$ denotes the query embedding of the $\left [ CLS \right ]$ token at Layer $l$ and Head $h$; $\mathbf{f}_{lh}^k(P)$ denotes the key embedding of Layer $l$ and Head $h$ for an image token at location $P$; $C$ is the number of channels for the query and key embedding.

The image tokens to mask are selected based on the scores $s_P$. We consider three strategies: 1) ``Low'' strategy. The image tokens with the lowest scores are discarded; 2) ``High'' strategy. The image tokens with the highest scores are discarded; 3) ``Mixed'' strategy. A portion of the maintained tokens are the ones with the highest scores, and the other maintained tokens are randomly selected. In the experiments, the ``low'' strategy performs best, and is set as the default in our method.

\paragraph{Generating scores using an EMA network}

To obtain the attentive masked images as described above, we need a network that can generate attention scores for all image areas. We introduce an exponential moving average (EMA) version of the CLIP visual encoder and apply this EMA network on the full image to generate the attention scores. Using the EMA network has the following merits: First, the attention scores can be computed online, without a need of an already trained network. Second, the EMA network constitutes another view of the image, so other effective auxiliary tasks such as BYOL~\cite{JeanBastienGrill2020BootstrapYO} can be introduced with little overhead. Third, the EMA network only involves the inference phase during training, which can save $2/3$ of the computational cost compared to using a trainable network. 

In order to further improve efficiency and effectiveness, we propose the following two techniques.

\paragraph{Efficient EMA computation with reduced image resolution} 
The EMA network is primarily used for attention score computation to select image tokens to maintain. It does not need to be very accurate. We thus propose to perform the EMA computation at reduced image resolution, for example, using half the original resolution. By reducing the resolution, the EMA computation cost is saved by more than $4\times$.

Overall, the EMA computation will correspond to be $1/12$ of a regular CLIP visual encoder, which is efficient.

\paragraph{Shared EMA score map for multiple masked views}

We also consider adding more masked image views for better results. A naive solution would be performing one time of the EMA computation for each view. However, this would waste computational cost. We propose using a shared EMA attention score maps for different image views.

Specifically, we first compute the minimum enclosing rectangle for multiple image views. Accordingly, we crop and resize the original image, where the EMA network is applied to produce an attention score map. The selection scores for each image view are computed from this shared attention score map via bilinear interpolation, and with the selection scores, the masking is performed using the “low” strategy described above. Please see the process in Figure~\ref{figure:attn}.

\subsection{Overall Framework}
\label{sec:overall}

\begin{figure}
\centering 
\includegraphics[width=\columnwidth]{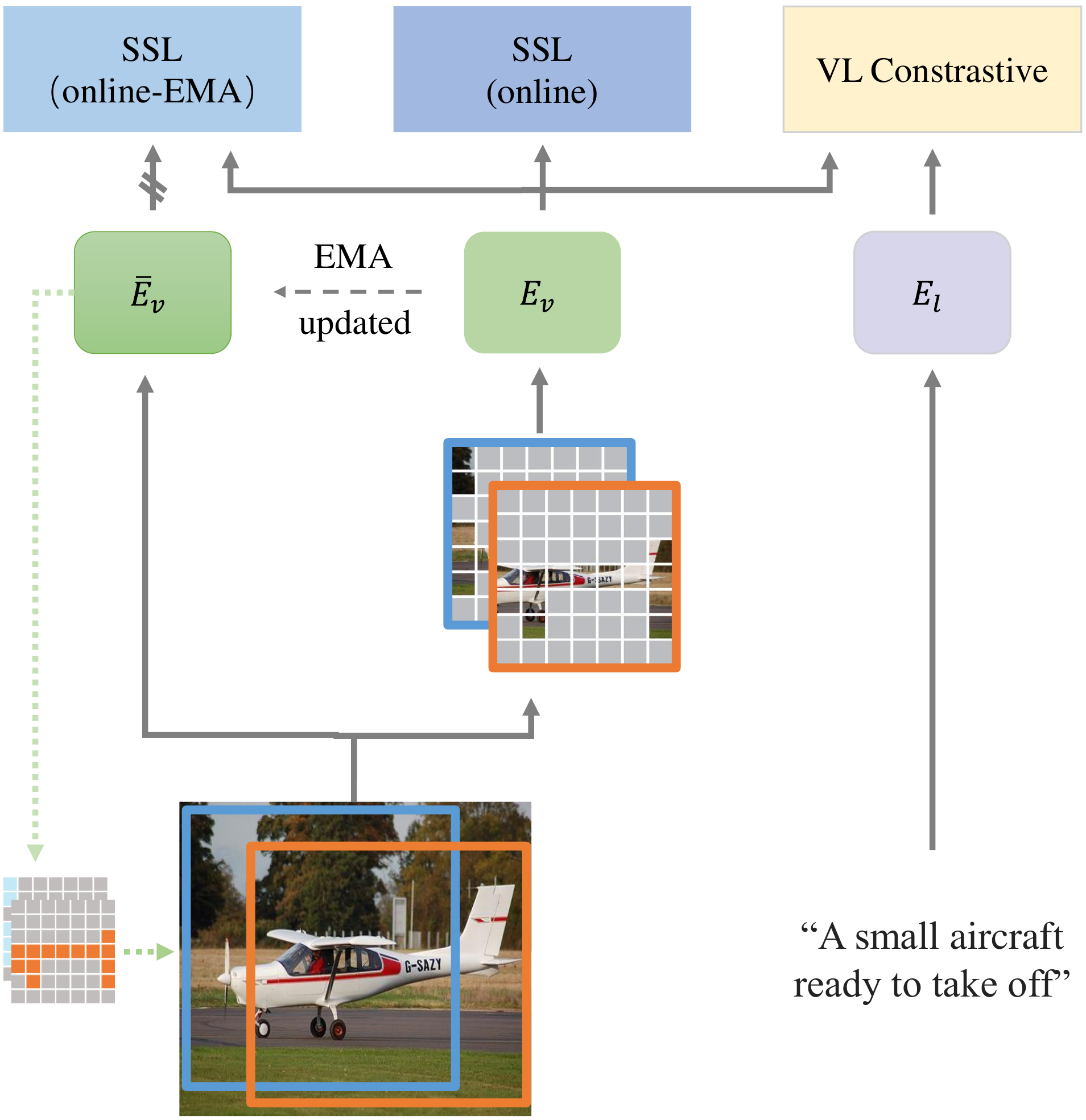} 
\caption{The architecture diagram of A-CLIP. We use masked multi-view input, where each view is computed independently for the vision language(VL) contrastive loss and then averaged. Between multiple online views we introduce auxiliary image self-supervised learning(SSL) to enrich  features. An EMA update vision encoder is used to help us generate the attentive mask, which is not involved in the gradient update. Since EMA encoder uses unmasked images, we design an self-distillation based online-EMA SSL task to learn the output distribution of the complete image.}  %
\label{figure:arch} 
\end{figure}

Based on attentive masking, we present an attentive mask CLIP method, dubbed A-CLIP. The overall framework is illustrated in Figure~\ref{figure:arch}. In the following, we describe several key designs and implementations of the A-CLIP framework.

\paragraph{Multiple masked views} The A-CLIP framework can take multiple masked views for better results. To make different number of views with roughly the same complexity, we set the token number for each masked view as $N/k$ ($N$ and k are the number of tokens and masked views respectively). The final VL loss is the average of all losses between each masked view and the alt-text. In the experiments, $k=2,3$ perform best (see Table~\ref{tab:views}), and $k=2$ is set as default.
\paragraph{Auxiliary self-supervised tasks}
Our framework can conveniently add several auxiliary self-supervised tasks. We consider two tasks: 1) Online-to-EMA contastive task. We adopt the BYOL instantiation~\cite{JeanBastienGrill2020BootstrapYO} for this task, which encourages the feature of a masked view to be the same as the EMA feature. 2) Online-to-online contrastive tasks. When multiple masked views are used, we can introduce contrastive losses between the masked views. We use the SimCLR~\cite{Simclr} or SimSiam~\cite{XinleiChen2020ExploringSS} instantiation for this task. Table~\ref{tab:ablation_distill} shows that both auxiliary self-supervised learning tasks can improve the zero-shot and retrieval accuracy. 


\paragraph{An A-CLIP-eff variant} For A-CLIP, we use the full image resolution as the input to the EMA network. We also introduce a variant called A-CLIP-eff, which uses half the resolution for EMA input. The new variant is more efficient than the A-CLIP variant, while being marginally worse in accuracy. Specifically, A-CLIP-eff is $0.86\times$ than that of the original CLIP model in training cost, much faster than the A-CLIP variant ($1.16\times$ of the original CLIP model), and slightly worse than A-CLIP in terms of zero-shot image classification and multi-modal retrieval accuracy (see Table~\ref{tab:main_results}).

\section{Experiments}
\subsection{Implementation Details}
\paragraph{Dataset and augmentation}
We train our model on a 15M subset of YFCC100M~\cite{BartThomee2022YFCC100MTN} filtered by Radford et al. ~\cite{AlecRadford2021LearningTV}. The text data in the subset consists of English-only titles and descriptions. 
During training, we randomly sample a valid caption (e.g., title or description) for each image, as in SLIP \cite{NormanMu2022SLIPSM}. 
The data augmentation is also similar to SLIP. Images in the online branches are randomly resized and cropped to between 50\% and 100\% of the original image size. For the EMA encoder, we choose a larger randomly cropped sub-image than online views to ensure we can get attentive scores by bilinear sampling. 
When adding auxiliary self-supervised learning tasks such as SimCLR or SimSiam, we used a classic combination of color jitter, grayscale, solarize and blur. 


\paragraph{Architecture and training configurations} We employ the ViT-B/16\cite{Vit} architecture as our visual encoder, and a 12-layer, 512-width, and 8-head Transformer as our text encoder following CLIP. The input resolution for the image encoder is 224$\times$224, and the text is encoded into 77 tokens using a vocabulary of 49k tokens with necessary truncations or paddings. To ensure stable training, we use a fixed random patch projection layer to embed image patches, following MoCo v3~\cite{chen2021empirical}.

The AdamW optimizer with a learning rate of 5e-4 and a weight decay of 0.5 is used, with a batch size of 4,096. The built-in automatic mixed precision library in PyTorch is adopted for training in all experiments.
For the exponential moving average (EMA) model which generates the attentive mask, its momentum starts from 0.996 and gradually increases to 1 using a cosine scheduler during training, following \cite{JeanBastienGrill2020BootstrapYO}. For the A-CLIP-eff variant, we use a halved resolution image as the input of the EMA model, with a bi-cubic interpolation method to get the new position encodings for the lower-resolution images. 


\paragraph{Details of Evaluation Tasks}
For zero-shot retrieval, we perform text-to-image and image-to-text evaluation on two benchmarks: COCO ~\cite{TsungYiLin2014MicrosoftCC} and Flickr30k~\cite{flickr30k}.
Following~\cite{AlecRadford2021LearningTV}, we retrieve top-$k$ candidates using the similarity scores from the image and text encoder. We evaluate zero-shot transfer capacity of the models on various classification benchmarks, such as ImageNet \cite{OlgaRussakovsky2014ImageNetLS} and Caltech101 \cite{LiFeiFei2006OneshotLO}. Following~\cite{AlecRadford2021LearningTV}, we use the same prompt templates and class names for the evaluation.

\subsection{Main Results} \label{sec:main_result}

\begin{table}[t]
\centering
\resizebox{0.85\columnwidth}{!}{%
\subfloat[Comparison of different mask strategies.]{

\begin{tabular}{c|c|cc|cc} 
\toprule
\multirow{2}{*}{\begin{tabular}[c]{@{}c@{}}Mask Strategy \\(View $\times$ Ratio)\end{tabular}} & IN 1K                & \multicolumn{2}{c|}{Flickr30K}       & \multicolumn{2}{c}{MS COCO}    \\ 
\cline{2-6}
                                                                                                            & 0-shot               & I2T           & T2I                  & I2T           & T2I            \\ 
\hline
w/o mask                                                                                              & \multicolumn{1}{c}{} &               & \multicolumn{1}{c}{} &               &                \\ 
\hline
1$\times$100\%                                                                                 & 37.6                 & 51.4          & 32.6                 & 27.9          & 17.6           \\ 
\hline
+random mask                                                                                        & \multicolumn{1}{c}{} &               & \multicolumn{1}{c}{} &               &                \\ 
\hline
1$\times$50\%                                                                                               & 35.0                 & 48.8          & 32.5                 & 28.9          & 16.6           \\ 
2$\times$50\%                                                                                               & 38.0                 & 54.6          & 34.4                 & 31.1          & 18.7           \\ 
\hline
+attentive mask                                                                                    & \multicolumn{1}{c}{} &               & \multicolumn{1}{c}{} &               &                \\ 
\hline
1$\times$50\%                                                                                               & 39.5                 & 57.6          & 36.6                 & 34.2          & 19.8           \\ 
\rowcolor[rgb]{0.925,0.957,1} 2$\times$50\%                                                                        & \textbf{41.3}        & \textbf{59.3} & \textbf{38.4}        & \textbf{35.1} & \textbf{21.3}  \\ 
\bottomrule
\end{tabular}
\label{tab:mask_ablation}
}
}
\quad
\resizebox{0.75\columnwidth}{!}{%
\subfloat[Ablations for adding attentive mask to CLIP training.]{
\begin{tabular}{c|c|cc|cc} 
\toprule
\multirow{2}{*}{Methods}           & IN 1K                & \multicolumn{2}{c|}{Flickr30K}       & \multicolumn{2}{c}{MS COCO}    \\ 
\cline{2-6}
                                   & 0-shot               & I2T           & T2I                  & I2T           & T2I            \\ 
\hline
Selection                          & \multicolumn{1}{c}{} &               & \multicolumn{1}{c}{} &               &                \\ 
\hline
high                               & 28.5                 & 42.6          & 29.0                 & 23.5          & 13.6           \\ 
mix                                & 40.4                 & \textbf{59.8} & 37.6                 & 34.9          & 20.9           \\ 
\rowcolor[rgb]{0.925,0.957,1} low  & \textbf{41.3}        & 59.3          & \textbf{38.4}        & \textbf{35.1} & \textbf{21.3}  \\ 
\hline
Patch Size                         & \multicolumn{1}{c}{} &               & \multicolumn{1}{c}{} &               &                \\ 
\hline
16                                 & 40.8                 & \textbf{61.4} & 37.6                 & \textbf{35.1}          & 20.5           \\ 
\rowcolor[rgb]{0.925,0.957,1} 32   & \textbf{41.3}        & 59.3          & \textbf{38.4}        & \textbf{35.1} & \textbf{21.3}  \\ 
\hline
Layers                             & \multicolumn{1}{c}{} &               & \multicolumn{1}{c}{} &               &                \\ 
\hline
last                               & 40.4                 & \textbf{59.4} & 36.8                 & 34.9          & 20.0           \\ 
\rowcolor[rgb]{0.925,0.957,1} all                                & \textbf{41.3}        & 59.3          & \textbf{38.4}        & \textbf{35.1} & \textbf{21.3}  \\ 
\hline
EMA                                & \multicolumn{1}{c}{} &               & \multicolumn{1}{c}{} &               &                \\ 
\hline
eff                                & 41.0                 & 56.8 & 37.5                 & \textbf{35.1}          & 20.4           \\ 
full & \textbf{41.3}        & \textbf{59.3}          & \textbf{38.4}        & \textbf{35.1} & \textbf{21.3}  \\
\bottomrule
\end{tabular}
\label{tab:attn_mask_ablation}
}
}
\caption{Comparison of different mask strategies and ablation experiments of attentive mask. The indigo background highlights our default setting.}
\label{tab:attn_ablation}
\end{table}

\paragraph{CLIP with random masking} 
This is the first work to analyze the impact of mask on the zero-shot performance of CLIP. 
There are two potential effects of mask training on CLIP. One is that it may cause ambiguity or meaninglessness in text and image matching, and the other is that the masked input of the pre-training process differs significantly from the full image during zero-shot evaluation. Our analysis in Table~\ref{tab:mask_ablation} reveals that randomly masking 50\% of image tokens in the 1$\times$50\% setting leads to a drop of -2.6\% in ImageNet-1K zero-shot accuracy compared to the original CLIP model, as well as -2.6/-0.1, +1.0/-1.0 I2T/T2I retrieval accuracy on Flickr30K and MS COCO. 

However, the comparison may be unfair due to lower training costs with mask training. To address this, we use two-views for the random masking method, which is still 20\% more efficient than the original CLIP model due to the superlinear computation cost reduction for the self-attention layers. The two-view version is now comparable in zero-shot accuracy to the one using full images (38.0\% vs. 37.6\%), and has shown +3.2/+1.8, +3.2/+1.1 retrieval accuracy improvements on Flickr30K and MS COCO. We believe that in the 2$\times$50\% case, two independent random crop samples increase the richness of the input visual signal, alleviate the side effects of image and text description mismatch caused by random mask, and gain the benefits of mask as an effective data augmentation at the same time.


\paragraph{The effects of attentive mask} 
As described in \Cref{sec:attn_mask}, we developed attentive mask strategies for CLIP training that allow us to selectively sample tokens based on their potential relevance to the text, rather than random masking. 

In the 1$\times$50\% setting, attentive mask improved performance by +4.5\%, +8.8/+4.1 and +5.3/+3.2 compared to random mask in ImageNet-1K, Flickr30K and MS COCO I2T/T2I. In the 2$\times$50\% setting, improvements to random mask were +4.7\%, +4.7/+4.0 and +4.0/+2.6. It can be found that the attentive mask solves the non-negligible performance degradation caused by random mask in single masked view setting, and is even much higher than full token CLIP. These results demonstrated that attentive mask can effectively integrate mask learning into CLIP, providing the benefits of data augmentation without introducing bias to the vision language contrastive task.

We also conducted experiments on different selection strategies, as shown in Table~\ref{tab:attn_mask_ablation}. The \textit{Low} strategy, which retained the top 50\% of the most relevant tokens, achieving the best results. It increased ImageNet-1K zero-shot performance by +3.7\%, and improved Flickr30K and MS COCO I2T/T2I retrieval accuracy by +7.9/+5.8 and +7.2/+3.7. This indicates that attentive selection can provide a more informative and valuable input distribution. In contrast, the \textit{High} strategy, which masks off the most relevant 50\% of tokens and is a good strategy for AttnMask \cite{attentive_mask} in MIM, resulted in a drastic performance collapse, while the \textit{Mix} strategy, which combines \textit{Low} with 25\% random selection, did not match the performance of \textit{Low}. These results suggest that for CLIP's VL contrastive task, only tokens with greater relevance are needed, which is consistent with the assumption of MaskCLIP \cite{XiaoyiDong2022MaskCLIPMS} that CLIP's training mainly relates to the language-described region and plays a low role.

Although masking brings efficiency gains, the additional cost of attentive selection is the need to pre-infer with the EMA encoder. Although back-propagation is not required, this takes close to 30\% of the training time. To further reduce the cost, we use half-resolution images as input to the EMA encoder, which reduces the cost to 5\%. As shown in the results of EMA-eff in Table~\ref{tab:attn_mask_ablation}, this step achieves competitive results compared to EMA-full and resulted in a +3.4\%, +5.4/+4.9, +7.2/+2.8 improvement on ImageNet-1K zero-shot classification, Flickr30K and MS COCO I2T/T2I retrieval, respectively.

\begin{table}[]
\centering
\resizebox{0.9\columnwidth}{!}{%
\begin{tabular}{c|c|cc|cc} 
\toprule
\multirow{2}{*}{Methods} & IN 1K                & \multicolumn{2}{c|}{Flickr30K}       & \multicolumn{2}{c}{MS COCO}    \\ 
\cline{2-6}
                         & 0-shot               & I2T           & T2I                  & I2T           & T2I            \\ 
\hline
CLIP                     & \multicolumn{1}{c}{} &               & \multicolumn{1}{c}{} &               &                \\ 
\hline
plain                    & 37.6                 & 51.4          & 32.6                 & 27.9          & 17.6           \\ 
+SimCLR(SLIP)            & 42.8                 & 57.2          & 41.2                 & 33.6          & 21.9           \\ 
+MAE(MaskCLIP)           & 42.7                 & 60.0          & 38.8                 & 34.1          & 21.2           \\ 
\hline
A-CLIP                   & \multicolumn{1}{c}{} &               & \multicolumn{1}{c}{} &               &                \\ 
\hline
plain                    & 41.3                 & 59.3          & 38.4                 & 35.1          & 21.3           \\ 
+SimCLR                  & 42.8                 & 63.6          & 41.0                 & 36.0          & 22.6           \\ 
\rowcolor[rgb]{0.925,0.957,1}+SimCLR+BYOL             & \textbf{43.9}        & 62.7          & \textbf{42.1}        & 38.0          & 23.2           \\ 
+SimSiam                 & 43.1                 & 62.5          & 41.3                 & 37.6          & 22.6           \\ 
+SimSiam+BYOL            & 43.4                 & \textbf{64.1} & 41.5                 & \textbf{38.1} & \textbf{23.3}  \\
\bottomrule
\end{tabular}
}
\caption{Ablation for auxiliary self-supervised tasks.}
\label{tab:ablation_distill}
\end{table} 

\begin{table}
\centering
\resizebox{0.85\columnwidth}{!}{%
\begin{tabular}{c|c|cc|cc} 
\toprule
\multirow{2}{*}{Methods}                    & IN 1K         & \multicolumn{2}{c|}{Flickr30K} & \multicolumn{2}{c}{MS COCO}    \\ 
\cline{2-6}
                                            & 0-shot        & I2T           & T2I            & I2T           & T2I            \\ 
\hline
CLIP(25ep)                                  & 37.6          & 51.4          & 32.6           & 27.9          & 17.6           \\
SLIP(25ep)                                  & 42.8          & 57.2          & 41.2           & 33.6          & 21.9           \\
\rowcolor[rgb]{0.925,0.957,1} A-CLIP(25ep)  & \textbf{43.9} & \textbf{62.7} & \textbf{42.1}  & \textbf{38.0} & \textbf{23.2}  \\ 
\hline
CLIP(50ep)                                  & 39.4          & 53.9          & 35.8           & 30.2          & 19.2           \\
SLIP(50ep)                                  & 44.1          & 60.6          & 41.1           & 33.2          & 22.3           \\
\rowcolor[rgb]{0.925,0.957,1} A-CLIP(50ep)  & \textbf{46.3} & \textbf{66.7} & \textbf{43.2}  & \textbf{39.8} & \textbf{24.4}  \\ 
\hline
CLIP(100ep)                                 & 42.7          & 61.0          & 37.9           & 34.4          & 20.9           \\
SLIP(100ep)                                 & 45.0          & 59.3          & 41.4           & 34.6          & 22.7           \\
\rowcolor[rgb]{0.925,0.957,1} A-CLIP(100ep) & \textbf{48.0} & \textbf{66.3} & \textbf{45.7}  & \textbf{40.7} & \textbf{25.1}  \\ 
\hline
CLIP(VIT-L)                                 & 40.4 & 51.4          & 35.2           & 28.9          & 18.5           \\
SLIP(VIT-L)                                 & 46.2 & 60.6          & 43.7           & 35.3          & 23.5           \\
\rowcolor[rgb]{0.925,0.957,1} A-CLIP(VIT-L) & \textbf{48.9}                            & \textbf{64.1} & \textbf{48.2}  & \textbf{39.1} & \textbf{26.9}  \\
\bottomrule
\end{tabular}
}
\caption{The results using longer training schedulers and bigger model size.}
\label{tab:epochs}
\end{table}

\begin{table}[]
\centering
\resizebox{0.85\columnwidth}{!}{%
\begin{tabular}{c|c|c|c|c} 
\toprule
\multirow{2}{*}{Methods} & \multicolumn{4}{c}{tokens per view $\times$ k views}            \\ 
\cline{2-5}
                         & 196 $\times$ 1 & 98 $\times$ 2 & 65 $\times$ 3 & 48 $\times$ 4  \\ 
\hline
+attentive mask          & 37.6           & \textbf{41.3} & \textbf{41.3} & 38.9           \\
\bottomrule
\end{tabular}
}
\caption{Effects of more attentive mask views. In this table, we tried more views, i.e., 3 and 4 views,  while keeping the total image tokens of multiple views to be the same (196 tokens) so as to keep the computation overheads roughly the same. Results evaluated by zero-shot classification accuracy on ImageNet-1K validation set.}
\label{tab:views}
\end{table}

\begin{table*}[]
\centering
\setlength{\aboverulesep}{0pt}
\setlength{\belowrulesep}{0pt}
\setlength{\tabcolsep}{2pt}
\linespread{1}
\scriptsize
\scalebox{1.0}{
\begin{tabular}{c|c|cccccccccccccccccccccccccc|c} 
\toprule
\rotatebox[origin=lb]{90}{\smash{Epochs}} &
  \rotatebox[origin=lb]{90}{\smash{Methods}} &
  \rotatebox[origin=lb]{90}{\smash{Food-101}} & \rotatebox[origin=lb]{90}{\smash{CIFAR-10}} & \rotatebox[origin=lb]{90}{\smash{CIFAR-100}} & \rotatebox[origin=lb]{90}{\smash{CUB}} & \rotatebox[origin=lb]{90}{\smash{SUN397}} &
\rotatebox[origin=lb]{90}{\smash{Cars}} & \rotatebox[origin=lb]{90}{\smash{Aircraft}} & \rotatebox[origin=lb]{90}{\smash{DTD}} & \rotatebox[origin=lb]{90}{\smash{Pets}} & \rotatebox[origin=lb]{90}{\smash{Caltech-101}} &
\rotatebox[origin=lb]{90}{\smash{Flowers}} & \rotatebox[origin=lb]{90}{\smash{MNIST}} & \rotatebox[origin=lb]{90}{\smash{FER-2013}} & \rotatebox[origin=lb]{90}{\smash{STL-10}} & \rotatebox[origin=lb]{90}{\smash{EuroSAT}} &
\rotatebox[origin=lb]{90}{\smash{RESISC45}} & \rotatebox[origin=lb]{90}{\smash{GTSRB}} & \rotatebox[origin=lb]{90}{\smash{KITTI}} & \rotatebox[origin=lb]{90}{\smash{Country211}} & \rotatebox[origin=lb]{90}{\smash{PCAM}} &
\rotatebox[origin=lb]{90}{\smash{UCF101}} & \rotatebox[origin=lb]{90}{\smash{Kinetics700}} & \rotatebox[origin=lb]{90}{\smash{CLEVR}} & \rotatebox[origin=lb]{90}{\smash{HatefulMemes}} & \rotatebox[origin=lb]{90}{\smash{SST2}} &
\rotatebox[origin=lb]{90}{\smash{ImageNet}} & \rotatebox[origin=lb]{90}{\smash{Average}} \\ \hline
\multirow{4}{*}{25} & CLIP & 50.6 & 66.0 & 34.5 & 38.8 & 51.1 & 4.0 & 5.4 & 21.2 & 28.5 & 60.9 & 53.3 & 8.4 & 17.3 & 90.5 & 30.2 & 21.5 & 6.1 & 35.1 & 10.5 & 53.5 & 28.5 & 22.1 & 10.8 & 52.4 & 50.7 & 37.6 & 34.2 \\
 & SLIP & 59.5 & 78.6 & 45.2 & 38.7 & 53.4 & \textbf{5.4} & 5.7 & \textbf{26.1} & 31.1 & 71.0 & 56.6 & 9.8 & 19.6 & 94.4 & 20.3 & 28.9 & \textbf{14.5} & 34.0 & 11.6 & \textbf{55.4} & 37.7 & 26.9 & \textbf{17.5} & \textbf{52.8} & \textbf{51.1} & 42.8 & 38.0 \\
 & MaskCLIP & \textbf{60.6} & 70.1 & 41.6 & \textbf{43.3} & 54.0 & 4.9 & \textbf{8.2} & 25.5 & \textbf{36.8} & 68.9 & 53.6 & \textbf{11.2} & 22.4 & 93.9 & \textbf{35.1} & 24.8 & 10.1 & 30.5 & 12.5 & 51.2 & 37.0 & 28.1 & 12.9 & \textbf{52.8} & 50.0 & 42.7 & 37.8 \\
 & {\cellcolor[rgb]{0.925,0.957,1}}A-CLIP & {\cellcolor[rgb]{0.925,0.957,1}}58.3 & {\cellcolor[rgb]{0.925,0.957,1}}\textbf{82.8} & {\cellcolor[rgb]{0.925,0.957,1}}\textbf{51.0} & {\cellcolor[rgb]{0.925,0.957,1}}43.0 & {\cellcolor[rgb]{0.925,0.957,1}}\textbf{57.0} & {\cellcolor[rgb]{0.925,0.957,1}}\textbf{5.4} & {\cellcolor[rgb]{0.925,0.957,1}}7.6 & {\cellcolor[rgb]{0.925,0.957,1}}26.0 & {\cellcolor[rgb]{0.925,0.957,1}}32.0 & {\cellcolor[rgb]{0.925,0.957,1}}\textbf{71.6} & {\cellcolor[rgb]{0.925,0.957,1}}\textbf{57.7} & {\cellcolor[rgb]{0.925,0.957,1}}9.8 & {\cellcolor[rgb]{0.925,0.957,1}}\textbf{29.7} & {\cellcolor[rgb]{0.925,0.957,1}}\textbf{95.4} & {\cellcolor[rgb]{0.925,0.957,1}}29.3 & {\cellcolor[rgb]{0.925,0.957,1}}\textbf{30.3} & {\cellcolor[rgb]{0.925,0.957,1}}13.1 & {\cellcolor[rgb]{0.925,0.957,1}}\textbf{35.2} & {\cellcolor[rgb]{0.925,0.957,1}}\textbf{13.5} & {\cellcolor[rgb]{0.925,0.957,1}}51.6 & {\cellcolor[rgb]{0.925,0.957,1}}\textbf{38.3} & {\cellcolor[rgb]{0.925,0.957,1}}\textbf{29.6} & {\cellcolor[rgb]{0.925,0.957,1}}14.1 & {\cellcolor[rgb]{0.925,0.957,1}}\textbf{52.8} & {\cellcolor[rgb]{0.925,0.957,1}}49.9 & {\cellcolor[rgb]{0.925,0.957,1}}\textbf{43.9} & {\cellcolor[rgb]{0.925,0.957,1}}\textbf{39.6} \\ 
\hline
\multirow{3}{*}{50} & CLIP & 55.2 & 77.0 & 43.8 & 38.9 & 49.0 & 4.7 & 6.3 & 23.5 & 27.2 & 63.5 & 56.2 & \textbf{12.5} & 30.2 & 92.1 & 21.0 & 31.9 & 7.4 & 33.6 & 10.9 & 50.8 & 35.5 & 24.8 & 14.0 & 49.9 & 50.1 & 39.4 & 36.5 \\
 & SLIP & 61.9 & 76.8 & 48.9 & 39.2 & 54.8 & 7.3 & 9.0 & \textbf{29.8} & 31.9 & \textbf{75.0} & 57.7 & 9.8 & 24.9 & \textbf{95.6} & 37.8 & 32.5 & 9.0 & \textbf{35.1} & 12.7 & 54.4 & 41.1 & 30.3 & 13.8 & 49.5 & 49.9 & 44.1 & 39.7 \\
 & {\cellcolor[rgb]{0.925,0.957,1}}A-CLIP & {\cellcolor[rgb]{0.925,0.957,1}}\textbf{62.2} & {\cellcolor[rgb]{0.925,0.957,1}}\textbf{81.5} & {\cellcolor[rgb]{0.925,0.957,1}}\textbf{53.7} & {\cellcolor[rgb]{0.925,0.957,1}}\textbf{48.2} & {\cellcolor[rgb]{0.925,0.957,1}}\textbf{58.7} & {\cellcolor[rgb]{0.925,0.957,1}}\textbf{8.3} & {\cellcolor[rgb]{0.925,0.957,1}}\textbf{10.2} & {\cellcolor[rgb]{0.925,0.957,1}}27.7 & {\cellcolor[rgb]{0.925,0.957,1}}\textbf{40.5} & {\cellcolor[rgb]{0.925,0.957,1}}73.3 & {\cellcolor[rgb]{0.925,0.957,1}}\textbf{61.0} & {\cellcolor[rgb]{0.925,0.957,1}}11.3 & {\cellcolor[rgb]{0.925,0.957,1}}\textbf{32.9} & {\cellcolor[rgb]{0.925,0.957,1}}95.5 & {\cellcolor[rgb]{0.925,0.957,1}}\textbf{39.7} & {\cellcolor[rgb]{0.925,0.957,1}}\textbf{37.5} & {\cellcolor[rgb]{0.925,0.957,1}}\textbf{9.4} & {\cellcolor[rgb]{0.925,0.957,1}}23.3 & {\cellcolor[rgb]{0.925,0.957,1}}\textbf{14.4} & {\cellcolor[rgb]{0.925,0.957,1}}\textbf{63.7} & {\cellcolor[rgb]{0.925,0.957,1}}\textbf{42.5} & {\cellcolor[rgb]{0.925,0.957,1}}\textbf{31.6} & {\cellcolor[rgb]{0.925,0.957,1}}\textbf{19.6} & {\cellcolor[rgb]{0.925,0.957,1}}\textbf{50.8} & {\cellcolor[rgb]{0.925,0.957,1}}\textbf{52.3} & {\cellcolor[rgb]{0.925,0.957,1}}\textbf{46.3} & {\cellcolor[rgb]{0.925,0.957,1}}\textbf{42.2} \\ 
\hline
\multirow{3}{*}{100} & CLIP & 60.4 & 79.4 & 44.6 & 43.3 & 53.0 & 8.5 & 8.2 & 26.2 & 34.7 & 68.9 & 59.2 & 11.4 & 20.4 & 93.2 & 23.3 & 27.3 & 10.3 & 23.1 & 12.0 & 54.0 & 36.7 & 27.7 & 13.0 & 50.9 & 50.1 & 42.7 & 37.8 \\
 & SLIP & 63.0 & 83.1 & 50.4 & 43.0 & 52.0 & 8.3 & 8.3 & 26.2 & 34.0 & 74.6 & 61.1 & \textbf{16.1} & 32.4 & 95.1 & 22.6 & 28.5 & 10.5 & \textbf{34.8} & 11.5 & 52.1 & 37.3 & 28.3 & 13.7 & \textbf{55.2} & 49.9 & 45.0 & 39.9 \\
 & {\cellcolor[rgb]{0.925,0.957,1}}A-CLIP & {\cellcolor[rgb]{0.925,0.957,1}}\textbf{66.7} & {\cellcolor[rgb]{0.925,0.957,1}}\textbf{86.6} & {\cellcolor[rgb]{0.925,0.957,1}}\textbf{58.6} & {\cellcolor[rgb]{0.925,0.957,1}}\textbf{51.4} & {\cellcolor[rgb]{0.925,0.957,1}}\textbf{58.6} & {\cellcolor[rgb]{0.925,0.957,1}}\textbf{10.5} & {\cellcolor[rgb]{0.925,0.957,1}}\textbf{11.9} & {\cellcolor[rgb]{0.925,0.957,1}}\textbf{33.1} & {\cellcolor[rgb]{0.925,0.957,1}}\textbf{48.5} & {\cellcolor[rgb]{0.925,0.957,1}}\textbf{74.9} & {\cellcolor[rgb]{0.925,0.957,1}}\textbf{64.3} & {\cellcolor[rgb]{0.925,0.957,1}}7.8 & {\cellcolor[rgb]{0.925,0.957,1}}\textbf{31.2} & {\cellcolor[rgb]{0.925,0.957,1}}\textbf{96.7} & {\cellcolor[rgb]{0.925,0.957,1}}\textbf{35.6} & {\cellcolor[rgb]{0.925,0.957,1}}\textbf{35.8} & {\cellcolor[rgb]{0.925,0.957,1}}\textbf{12.9} & {\cellcolor[rgb]{0.925,0.957,1}}30.5 & {\cellcolor[rgb]{0.925,0.957,1}}\textbf{15.7} & {\cellcolor[rgb]{0.925,0.957,1}}\textbf{57.1} & {\cellcolor[rgb]{0.925,0.957,1}}\textbf{44.1} & {\cellcolor[rgb]{0.925,0.957,1}}\textbf{33.1} & {\cellcolor[rgb]{0.925,0.957,1}}\textbf{22.9} & {\cellcolor[rgb]{0.925,0.957,1}}52.7 & {\cellcolor[rgb]{0.925,0.957,1}}\textbf{50.7} & {\cellcolor[rgb]{0.925,0.957,1}}\textbf{48.1} & {\cellcolor[rgb]{0.925,0.957,1}}\textbf{43.8} \\
\bottomrule
\end{tabular}
}
\caption{Zero-shot evaluation on a variety of classification benchmarks. The \textit{Epochs} indicates the number of training rounds. A-CLIP significantly outperforms other methods at all epochs setting, both in terms of average accuracy and number of winning tracks of above 25 downstream tasks.}
\label{tab:downstream}
\end{table*}

\paragraph{A-CLIP provides an efficient paradigm for combining SSL with CLIP}

SLIP~\cite{NormanMu2022SLIPSM} and MaskCLIP~\cite{XiaoyiDong2022MaskCLIPMS} both add image self-supervised learning (SSL) to CLIP using a separate branch, whereas our proposed A-CLIP integrates different SSL tasks naturally and efficiently through masked multi-view input. In \Cref{tab:ablation_distill}, with the same SimCLR task added, A-CLIP can achieve comparable ImageNet \cite{OlgaRussakovsky2014ImageNetLS} zero-shot performance to SLIP \cite{NormanMu2022SLIPSM}, +6.4/-0.2 in Flickr30K retrieval and +2.4/+0.7 in MS COCO retrieval, but the reduced branches accelerate the training time by 2.3$\times$. We tried SimSiam without using negative samples as a SSL task and had similar performance gains, demonstrating that our approach is not dependent on specific SSL tasks.

Online-EMA SSL can further improve performance. To obtain the attentive mask through the EMA encoder, we use a self-distillation based BYOL head to learn the output distribution of the full image. Table \ref{tab:ablation_distill} shows that Online-EMA task BYOL improves performance in both tasks, and it is complementary to online SSL.

\paragraph{Zero-shot performance on 25 benchmarks}
We tested the zero-shot classification performance of the proposed A-CLIP approach on a larger set of validation benchmarks, which included 25 classification tasks, following SLIP's \cite{NormanMu2022SLIPSM} evaluation setting. In testing, we also followed SLIP's prompts for evaluation for a fair comparison.

Table \ref{tab:downstream} reports the results under different training lengths. A-CLIP achieves +1.6\%, +2.5\%, and +3.9\% gains over the counterpart SLIP method regarding the average accuracy of this 25-dataset suite, using 25-epoch, 50-epoch, and 100-epoch training, respectively. Also, note that A-CLIP is 2.3$\times$ faster than SLIP. These results further prove the efficacy and efficiency of the proposed A-CLIP.
\paragraph{Comparison with other methods}
In \Cref{tab:main_results}, we compare the differences with other methods. Using ViT-B and YFCC-15M dataset, our approach achieves 43.9\% top-1 accuracy on ImageNet-1K zero-shot classification, as well as 62.7/42.1 and 38.0/23.2 I2T/T2I retrieval accuracy on Flickr30K and MS COCO, which are +1.1\% higher, +5.5/+0.9 higher, and +4.4/+1.3 higher than the SLIP method, respectively, while being 2.30$\times$ faster and requiring less GPU memory. A-CLIP-eff, an efficient version that is even 1.16$\times$ faster than the plain CLIP model, achieves significant gains of +5.3\%, +11.3/+8.0, and +9.5/+4.9 on these benchmarks over that of the plain CLIP model.

\subsection{Ablation Study and Analysis}

\paragraph{Effects of training longer}
\Cref{tab:epochs} ablates the effects of longer training for different frameworks. Using 2$\times$ and 4$\times$ training schedulers, i.e., 50 and 100 epochs, the accuracy gaps between the A-CLIP and SLIP models are larger on benchmarks than that using 25-epoch training: +1.1\%, \textbf{+2.2\%}, and \textbf{+3.0\%} gains ImageNet-1K zero-shot classification, and +4.4/+1.3, \textbf{+6.6/+2.1}, and \textbf{+6.1/+2.4} on MS COCO~\cite{TsungYiLin2014MicrosoftCC} I2T/T2I retrieval, by 25, 50, and 100 epochs training, respectively. On Fickr30K~\cite{flickr30k}, longer training also overall benefits more for A-CLIP: +5.5/+0.9, \textbf{+6.1/+2.1}, and \textbf{+7.0/+4.3} gains on I2T/T2I retrieval by 25, 50, and 100 epochs training, respectively.

We hypothesize that the attentive mask input play as strong augmentation to the input images, which can greatly alleviate the over-fitting issue and thus perform better when training is longer.

\paragraph{Effects of larger model size}
In Table~\ref{tab:epochs}, we have included a comparison with ViT-L/16 on 25 epochs setting. A-CLIP benefits more with larger model size.

\begin{figure*}[ht]
\centering 
\includegraphics[width=1.0\textwidth]{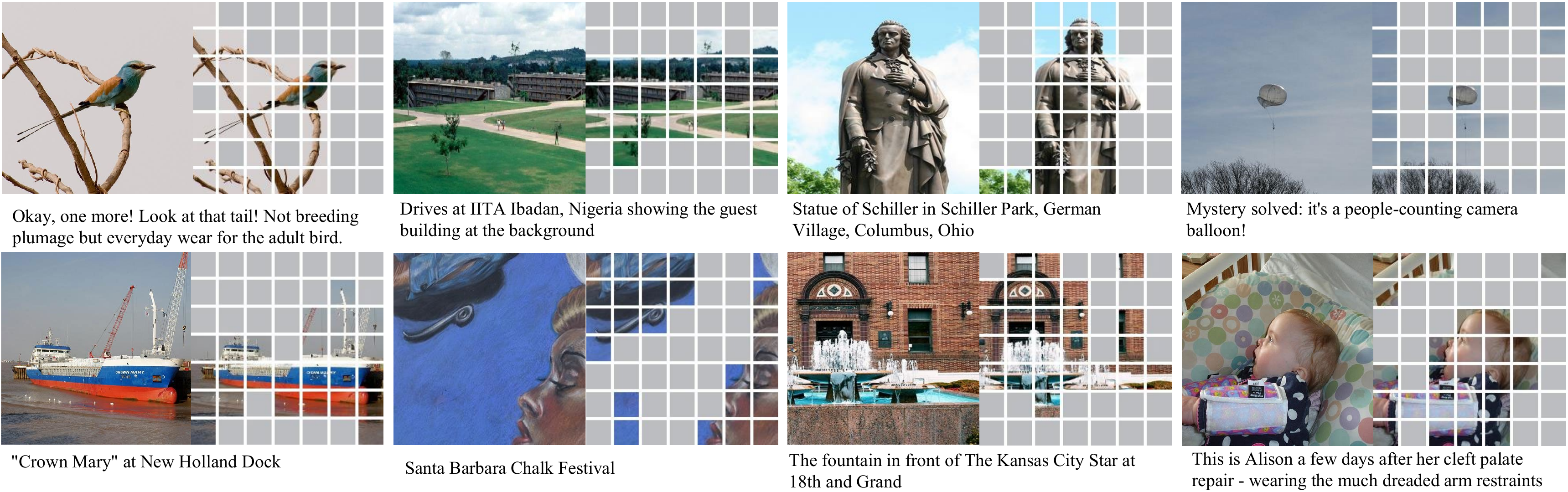} 
\caption{Visualization of attentive mask. Here we use a ViT-B16 model from A-CLIP's EMA vision encoder of to generate a mask with patch size of $32 \times 32$ and $50\%$ mask ratio. Attentive mask can be found to magically preserve the content of text descriptions and filter out redundant backgrounds. The image and text used are sampled from YFCC-100M~\cite{BartThomee2022YFCC100MTN}.} 
\label{figure:vis} 
\end{figure*}

\paragraph{Effects of more attentive mask views }
Our proposed A-CLIP is not limited to two views, and can be extended for multiple views such as 3 and 4. 
\Cref{tab:views} shows the results of using different views. In this comparison, we keep the total number of tokens the same for different $k$ views settings. $k=1$ means the the original CLIP model which keeps all image tokens. For other $k$, each view selects the most relevant $196/k$ tokens. So the computation overheads for different $k$ are roughly the same.

It can be seen that all $k=2,3,4$ performs significantly better than that of $k=1$. $k=2,3$ both perform the best, and are higher than $k=4$ (+2.4\%). 

\paragraph{Use alt-text to select the image patches?}
We tried using alt-text to select image regions, yielding high training but low evaluation accuracy due to information leakage. Using the text to directly select its corresponding image alters the distance between every positive pairs, rendering the contrastive learning trivial. To address this, we employ EMA attention weights in Eq.(\ref{eq:Attn}) as a form of regularization, effectively avoiding direct information leakage. 

\paragraph{Visualization }
We have done some visualization of the attentive mask. Figure~\ref{figure:vis} showes the original image and the two views inputted in A-CLIP. It can be found that we basically keep the significant part of the image. The token that is closer to the language semantics makes the input of A-CLIP effective and valuable.

\paragraph{Effects of masked patch size } 
Referring to the conclusion of SimMIM~\cite{ZhendaXie2021SimMIMAS}, we experimented with different mask patch sizes. As can be seen in \Cref{tab:attn_ablation}, it obtained a +0.5\% ImageNet-1K zero-shot improvement from change mask size to 32. Due to the continuity and redundancy of images, greater granularity of filtering may lead to better performance.
\paragraph{Using all layers works better for attentive token selection } 
\Cref{tab:attn_ablation} ablates the use of different layers for attentive token selection. Using all layers performs +0.9\% higher on ImageNet-1K zero-shot than that using the last layer.
\section{Conclusion}

In this paper, we present the attentive mask CLIP framework, or A-CLIP for short, which achieves more efficient and more effective CLIP training by introducing attentive masks for the image branch. Compared to a baseline method using random masking, the attentive mask approach maintains image tokens that are more relevant to the alt-text description and performs much better. The A-CLIP framework is also flexible in incorporating multiple masked views as well as several auxiliary self-supervised tasks, which can further boost efficacy. The attentive mask also serves as good data augmentation, which benefits more from longer training than the original CLIP method. 

Due to these advantages, the A-CLIP framework performs much better in both efficiency and efficacy than other CLIP improvements such as SLIP and MaskCLIP. An efficient variant of A-CLIP-eff is even more efficient than the original CLIP method, while being significantly better in terms of zero-shot and multi-modal retrieval accuracy.



{\small
\bibliographystyle{ieee_fullname}
\bibliography{main}
}
\renewcommand\thesection{\Alph{section}}
\renewcommand{\thetable}{A\arabic{table}}
\renewcommand{\thefigure}{A\arabic{figure}}
\appendix  

\begin{sidewaysfigure*}[]
\centering 
\includegraphics[width=1.0\textwidth]{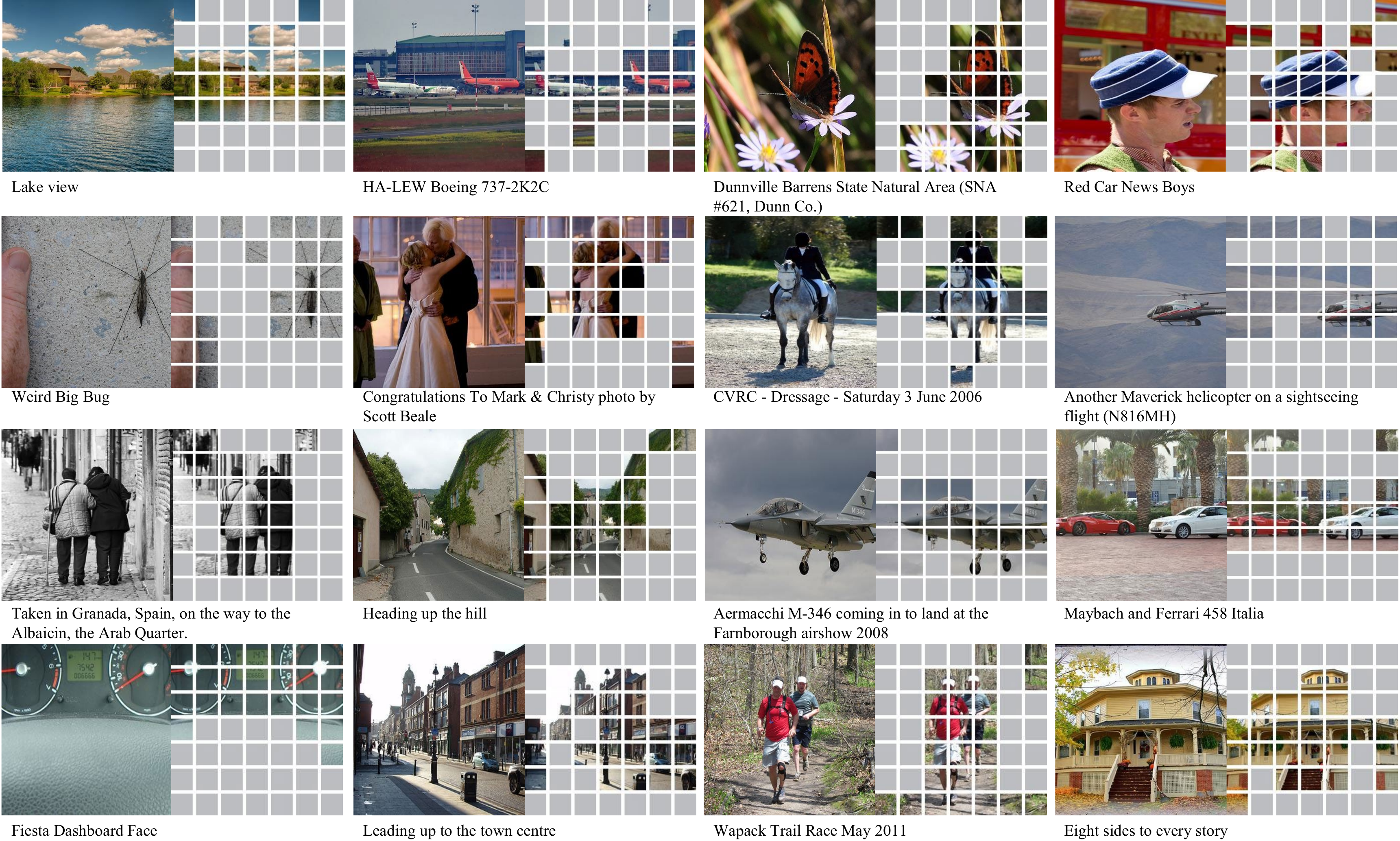}
\caption{More visualization of attentive mask. Here we use a ViT-B16 model from A-CLIP's EMA vision encoder of to generate a mask with patch size of $32 \times 32$ and $50\%$ mask ratio. The image and text used are sampled from YFCC-100M.} 
\label{figure:more_vis} 
\end{sidewaysfigure*}

\section{Additional Implementation Details}

\paragraph{Details of compared methods} 
For CLIP and SLIP, we use the publicly available model and checkpoints provided by SLIP\footnote{https://github.com/facebookresearch/SLIP}. 
We reproduced MaskCLIP ourselves since it is not open-sourced and does not report results on YFCC15M. We follow the hyperparameters in the MaskCLIP paper: we mask out 75\% of tokens in the mask self-distillation branch with a loss weight of 10.0; we start the moment of the EMA updated visual encoder from 0.999 and increase it linearly to 0.9999; we standardize the targets from the EMA encoder by a parameter-free Layer Norm; we set the learning rate to 5e-4, the batch size to 4,096, and the weight decay to 0.5.

We conduct all experiments on 4 nodes with 8 NVIDIA Tesla V100 GPUs each. We perform a speed test of different frameworks using a single node of 8 NVIDIA A100 GPUs to eliminate the effect of network conditions, which has been found to have very stable speed profiles for each framework.

\section{Effects of stronger data augmentation }

To combine more auxiliary self-superivise learning(SSL) tasks as mentioned in Section 3.2, we considered adding stronger data augmentation to learn robuster representation. In A-CLIP with or without SSL task, we tried with and without stronger data augmentation respectively. 

The results in Table \ref{tab:augs} demonstrate that the plain A-CLIP, which incorporates an attentive mask strategy into CLIP, benefits from the addition of the classic data augmentations of color+blur, leading to a +1.2\% boost in ImageNet-1K zero-shot top-1 accuracy. This suggests that using attentive masking as an efficient data augmentation does not conflict with traditional methods like color and blur. Regarding the addition of SIMCLR, the results show that it relies more heavily on stronger data augmentations, as its performance is weaker when only random crop is added. However, after incorporating the BYOL task, an online-EMA SSL, the model's performance steadily improves regardless of the strength of the data augmentations. We believe that this is because the BYOL task enables the model to learn the output distribution of the complete image from the EMA encoder.

\begin{table}[H]
\centering
\resizebox{0.9\columnwidth}{!}{%
\begin{tabular}{c|c|c} 
\toprule
\multirow{2}{*}{Methods}    & \multicolumn{2}{c}{IN 1K 0-shot}  \\ 
\cline{2-3}
                            & crop          & crop+color+blur   \\ 
\hline
+attentive mask             & 41.3          & 42.5              \\
+attentive mask+SimCLR      & 39.0          & 42.8              \\
+attentive mask+SimCLR+BYOL & \textbf{41.9} & \textbf{43.9}     \\
\bottomrule
\end{tabular}
}
\caption{Effects of different data augmentations for A-CLIP. Adding color+blur improves the performance of all settings, but more significantly after adding image self-supervised learning tasks.}
\label{tab:augs}
\end{table}

\section{Effects of using EMA inference }
A-CLIP uses an EMA vision encoder for inference in order to generate a stable attentive mask. In \Cref{tab:EMA}, we find that using EMA for evaluation leads to a stable performance gain, especially for mask training. 
Without mask, CLIP improves by +0.4\% on ImageNet 1K zero-shot classification using EMA; with attentive mask training, it improves by +1.3\%. We speculate that EMA alleviates the bias from the mask training.

\begin{table}[H]
\centering
\resizebox{0.65\columnwidth}{!}{%
\begin{tabular}{c|c|c} 
\toprule
\multirow{2}{*}{Methods} & \multicolumn{2}{c}{IN 1K 0-shot}     \\ 
\cline{2-3}
                         & online        & EMA                  \\ 
\hline
w/o mask                 & 37.6          & 38.0(+0.4)           \\
+random mask             & 38.0          & 39.1(+1.1)           \\
+attentive mask          & \textbf{40.0} & \textbf{41.3(+1.3)}  \\
\bottomrule
\end{tabular}
}
\caption{Performance comparison of using online and EMA vision encoders for evaluation on IN 1K 0-shot classification with different mask methods. EMA improves performance more significantly with mask training than without mask.}
\label{tab:EMA}
\end{table}

\section{More visualization results for attentive mask}
Figure~\ref{figure:more_vis} shows more visualization results of attentive mask for A-CLIP. It can be seen that attentive mask always retains the text-related areas, while the deleted areas are more redundant and non-text-related parts, which is what our motivation wants to achieve.
\end{document}